\documentclass[11pt,a4paper]{article}
\usepackage[margin=22mm]{geometry}
\usepackage[T1]{fontenc}
\usepackage{lmodern,microtype}
\usepackage{authblk}
\usepackage{amsmath,amssymb,graphicx,booktabs,array,tabularx}
\usepackage{xcolor}
\usepackage{subcaption,tikz}
\usetikzlibrary{arrows.meta,positioning,backgrounds}
\usepackage[section]{placeins}
\usepackage[numbers,sort&compress]{natbib}
\usepackage{xurl}
\usepackage[colorlinks=true,linkcolor=navy,citecolor=navy,urlcolor=navy]{hyperref}
\definecolor{navy}{HTML}{173D58}
\definecolor{teal}{HTML}{217D88}
\definecolor{amber}{HTML}{B77924}
\definecolor{coral}{HTML}{B14F4B}
\definecolor{ink}{HTML}{243441}
\graphicspath{{figures/}}
\newcommand{\code}[1]{\texttt{#1}}

\newcommand{\supp}{\hyperref[supp:start]{Supplementary Material}}
\hypersetup{pdftitle={Natural-Language to Nav2: Operational Policies for Costmap-Based Navigation}}
\newif\ifanonymous
\anonymousfalse 
\newcommand{\ProjectURL}{https://a56239637-dot.github.io/Natural-Language-to-Nav2-Project-Page/}

\title{\vspace{-1.4em}Natural-Language to Nav2:\\Operational Policies for Costmap-Based Navigation}
\ifanonymous
  \author{Anonymous authors}
  \hypersetup{pdfauthor={}}
\else
  \hypersetup{pdfauthor={Jiyeong Chae, Hyunkyo Seo, Heonjae Lee, Kyung-Joon Park}}

\author{Jiyeong Chae}
\author{Hyunkyo Seo}
\author{Heonjae Lee}
\author{Kyung-Joon Park}
\affil{Daegu Gyeongbuk Institute of Science and Technology (DGIST)\\
333 Techno Jungang-daero, Hyeonpung-eup, Dalseong-gun,\\
Daegu, 42988, Republic of Korea}
\newcommand{\authorinformation}{%
\vspace{-1.5em}
\begin{center}
\small Jiyeong Chae and Hyunkyo Seo contributed equally.\\
Corresponding author: Kyung-Joon Park (\href{mailto:kjp@dgist.ac.kr}{kjp@dgist.ac.kr}).
\end{center}}

\fi
\date{}
\begin{document}
\maketitle
\ifanonymous\else\authorinformation\fi
\begin{abstract}
Autonomous mobile robots support material transport in warehouses and industrial facilities, where navigation must adapt to changing operating conditions. Sensor observations reveal physical obstacles, but do not by themselves specify temporary access rules, route preferences, or exceptions known to operators. This paper presents a natural-language policy interface that brings this knowledge into ROS~2 Nav2. A local large language model converts instructions into JSON policies specifying destinations, spatial constraints, and temporal or event conditions. Deterministic checks govern mission acceptance, and three costmap adapters project accepted policies into the existing navigation stack without changing its planner or controller. On 150 researcher-authored English commands, four local models produce 107--142 correct accepted policies, compared with 44 for a rule-based parser, with 3--20 incorrect acceptances per model. Fault injection records rejection of eight endpoint/JSON faults and 29 validator-stage faults. Navigation experiments use correct task policies to evaluate keepouts, zone costs, person-dependent regions, and event-conditioned passage in simulation and on a TurtleBot3. For Tasks~1--3, mean travel times are 18.4--24.5\% lower in simulation and 23.1--35.0\% lower on the physical testbed than with the baseline lacking those policies. A native KeepoutFilter comparison and two-robot demonstrations examine basic keepout behavior and shared policy projection. The interface connects operator instructions to explicit navigation policies while retaining Nav2's planning and control components.
\end{abstract}
\noindent\textbf{Keywords:} mobile robot navigation; natural-language policies; large language models; ROS~2 Nav2; costmaps.
\section{Introduction}
Autonomous mobile robots (AMRs) support material transport in warehouses and industrial facilities~\citep{barros2021fulfillment,cheng2022dynamic}. Their navigation must account for changing workspace conditions as well as the destination of each delivery~\citep{pakkala2025design}. Some of these conditions are operating rules: an aisle may be reserved for loading, a corridor may be closed for a shift, or an emergency may permit an otherwise restricted passage. Sensors can detect an obstruction, but its expected duration and permitted exceptions require additional information. Operators need a way to make this knowledge available to the robot when planning a mission.

ROS~2 Nav2 provides planners and controllers for mobile navigation~\citep{macenski2020marathon}. Layered costmaps combine spatial information~\citep{lu2014layered}, and native keepout filters can receive updated masks during operation~\citep{nativekeepout}. These mechanisms provide a means to apply constraints once they have been expressed in a form the navigation stack can use. An instruction such as ``avoid Zone~A for the next thirty minutes'' also requires the system to identify the zone, retain the restriction, and determine when it expires. The problem addressed here is to connect such instructions to Nav2 with explicit rules for policy acceptance, mission replacement, and restriction lifetime.

\begin{figure}[htbp]
\centering
\includegraphics[width=\linewidth]{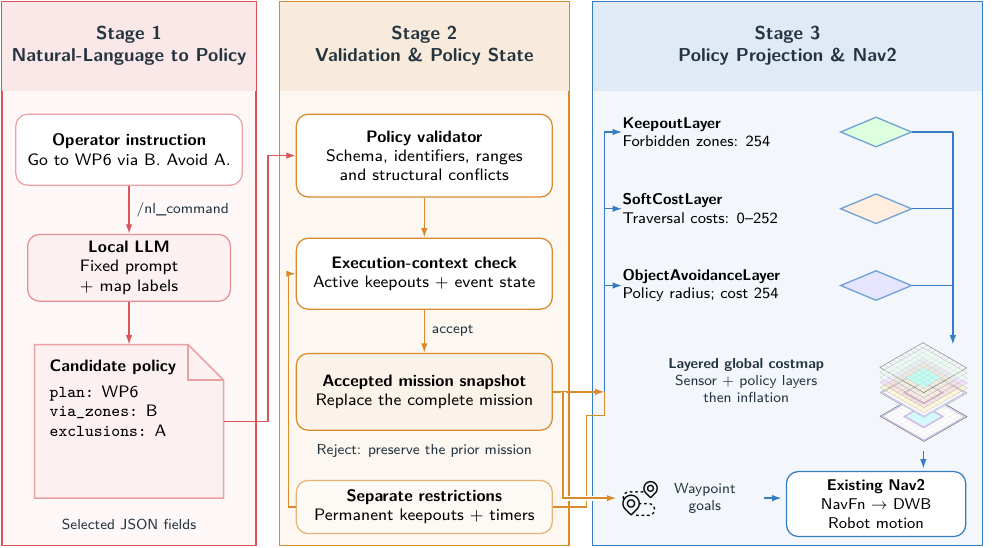}
\caption{The policy interface has three stages. The LLM receives an instruction and registered map labels. Parsing, validation, and execution-context checks govern mission acceptance. Accepted state supplies waypoint goals; policy state and separately maintained restrictions determine the cost contributions. Standard sensor layers, inflation, NavFn, and DWB remain in the navigation stack. The example shows selected policy fields.}
\label{fig:framework}
\end{figure}

We introduce a natural-language policy interface that represents destinations together with constraints on reaching them. A locally hosted large language model (LLM) translates each instruction into a structured policy. The bridge checks the complete candidate against the schema and execution context before accepting it. Accepted state supplies waypoint goals and inputs to three costmap adapters for keepouts, zone costs, and object-avoidance regions (Figure~\ref{fig:framework}). The same policy-derived updates can be sent to independent robot instances.

The LLM operates when a policy is submitted, outside the navigation control loop. NavFn computes the global path and DWB controls local motion. The intermediate policy makes the requested destinations, constraints, and conditions available for inspection. Structural acceptance and semantic correctness remain distinct: a candidate may satisfy the interface contract while naming a waypoint the operator did not request. The evaluation measures both.

The work makes three contributions. (1) A \textbf{policy schema} represents destinations together with spatial constraints and temporal or event conditions, providing a common interface between language interpretation and navigation. (2) \textbf{Transactional acceptance} checks the complete candidate against the schema and execution context before replacing the mission snapshot; rejection preserves the preceding mission, while retained restrictions are managed separately. (3) \textbf{Nav2 projection} translates accepted state into waypoint goals and keepout, soft-cost, and object-avoidance inputs for the existing navigation stack, including shared policy updates to independent robot instances.

The evaluation follows this architecture. A 150-command benchmark measures conversion accuracy and accepted errors, and fault injection examines rejection and state preservation. Simulation and physical trials examine route selection under task policies, including a separate comparison of event-conditioned passage permissions.

\clearpage
\section{Related Work}
\label{sec:related}
\paragraph{Language as a robot interface.}
ROSGPT maps natural-language requests to structured ROS~2 instructions using an application ontology and prompting~\citep{koubaa2023rosgpt}. Our interface extends the language-to-structured-command workflow with explicit management of mission state, restriction lifetimes, and event-conditioned permissions. Code as Policies generates programs that process perception outputs and invoke control APIs, including reactive and waypoint-based behaviors~\citep{liang2023code}.

LLM-A* uses language-model guidance in heuristic search~\citep{meng2024llm}, and 3P-LLM studies language-model-based path planning~\citep{latif20243p}. In our system, the model selects registered destinations and policy parameters; Nav2 computes geometric paths.

\paragraph{Context in costmaps.}
Layered costmaps combine spatial information from different sources~\citep{lu2014layered}. Dynamic-obstacle extensions include the ROS~2 layer approach of Cen Cheng et al.~\citep{cheng2022dynamic} and the velocity-obstacle costmap of Chen et al.~\citep{chen2022velocity}. Sani et al. use pixel-level image classification to revise local traversal costs for agricultural navigation~\citep{sani2024improving}. Our adapters project accepted operator policies into this costmap representation.

\paragraph{Nav2 and operational management.}
Nav2 separates planning and control from costmap generation~\citep{macenski2020marathon}. Its native KeepoutFilter accepts occupancy-grid masks and processes new masks at runtime~\citep{nativekeepout}. Table~\ref{tab:policy-comparison} compares this filter mechanism with the policy-management functions of our interface. The native/custom experiment isolates their common keepout function.

Industrial studies address fulfillment operations~\citep{barros2021fulfillment}, knowledge-based material handling~\citep{pakkala2025design}, and multi-AGV collision and deadlock resolution~\citep{zajac2021structural}. Our policy-sharing experiment distributes common restrictions to independent Nav2 instances.

\begin{table}[htbp]
\centering\small
\caption{Keepout filtering and policy management. The native column covers KeepoutFilter; Nav2 task orchestration lies outside that component.}
\label{tab:policy-comparison}
\setlength{\tabcolsep}{4pt}
\renewcommand{\arraystretch}{1.12}
\begin{tabularx}{\linewidth}{@{}>{\raggedright\arraybackslash}p{27mm}*{2}{>{\raggedright\arraybackslash}X}@{}}\toprule
Aspect & Native KeepoutFilter~\citep{nativekeepout} & Proposed policy interface \\\midrule
Policy validation & Filter metadata checks & Whole-candidate schema, conflict, and context checks \\\addlinespace[3pt]
Mission replacement & Outside filter scope & Replace mission snapshot; preserve separately managed restrictions \\\addlinespace[3pt]
Restriction lifetime & Retained mask; externally updated & Mission-scoped, permanent, and timed restrictions \\\addlinespace[3pt]
Event handling & External mask updates & Event-conditioned permissions with priorities \\\bottomrule
\end{tabularx}
\end{table}

\section{From an Instruction to a Navigation Policy}
\label{sec:approach}
The system treats an operating instruction as a proposed change to a mission. Figure~\ref{fig:framework} follows that change from language input to robot motion. Interpretation produces a candidate, acceptance establishes the new mission state, and projection converts the applicable rules into Nav2 inputs. All LLM inference runs on an on-premises workstation connected to the robots through the local network.

\subsection{A schema for destinations and operating rules}
An operator publishes an instruction on \code{/nl\_command}. A frozen local LLM receives the sentence, the registered zone and waypoint context, an output schema, and example conversions. No model weights are fine-tuned. Waypoint labels resolve to stored poses and zone labels resolve to map-frame geometry. This avoids asking the model to invent region coordinates.

Table~\ref{tab:schema} groups the policy fields by execution role. For example, ``Go to waypoint~6 via Zone~B, but do not use Zone~A'' requests a destination, a required transit zone, and a keepout. ``Set the cost of Zone~A to 200 and Zone~B to 160, then go to waypoint~4'' adds relative traversal costs. Temporal exclusions and conditional rules carry the lifetime or event scope that would otherwise be lost when reducing a command to a destination alone.

\begin{table}[htbp]
\caption{The JSON policy contract. A candidate may also explicitly reject an instruction using \code{decision} and \code{reason\_code}.}
\label{tab:schema}
\centering\small
\begin{tabularx}{\linewidth}{@{}p{46mm}X@{}}\toprule
Fields & Interpretation and execution \\\midrule
\code{plan}, \code{via\_zones} & Ordered destinations and required transit zones. The bridge resolves transit waypoints and builds the goal sequence. \\
\code{exclusions} & Zones to keep closed, with optional duration or absolute time interval. \\
\code{weights} & Integer zone costs for traversal preferences. \\
\code{dynamic\_object\_rules} & Object class and avoidance radius used around detections. \\
\code{conditional\_rules} & Zone actions selected by event/state conditions and priority. \\\bottomrule
\end{tabularx}
\end{table}

\subsection{Acceptance depends on the candidate and the current context}
An endpoint failure or malformed JSON is rejected before a policy object reaches the validator. A parsed candidate is checked for required structure, data types, registered identifiers, admissible values, and conflicts among fields. A zone cannot simultaneously be a required transit region and an exclusion. Conflicting costs for the same zone also invalidate the candidate. The entire update is rejected if a check fails. Valid fields are not applied selectively, and invalid numbers are not clipped.

The schema admits \code{person} and \code{forklift} object rules with finite radii from 0.1 to 10~m. Durations must be positive and no greater than 1,440~minutes; absolute intervals require ordered endpoints. Conditional rules use \code{default}, \code{fire\_alarm}, or \code{low\_battery} and the actions \code{forbid}, \code{allow}, or \code{allow\_shortest}. Default rules only forbid. Equal-priority conflicts, incompatible actions, and event priorities that fail to exceed a default prohibition are rejected. Keepout and soft-cost assignments to the same zone are checked for conflict.

A structurally admissible candidate can still conflict with a restriction already in force. The execution-context check tests required transit zones against effective keepouts, including permanent restrictions, retained timed restrictions, and applicable event permissions. It rejects an incompatible candidate without deleting the transit requirement. These checks establish structural and contextual consistency; they cannot detect every requirement omitted from the instruction.

Let $m$ denote the mission snapshot, $r$ the separately maintained restrictions, and $e$ the event state. Validation and context checking yield a normalized candidate $u^*=V(u;r,e)$ or rejection $\bot$. The commit operation is
\begin{equation}
 m^+=\begin{cases}u^*,&V(u;r,e)=u^*,\\m,&V(u;r,e)=\bot.\end{cases}
 \label{eq:commit}
\end{equation}
Acceptance replaces the complete mission snapshot within the bridge; downstream costmap and robot updates are asynchronous. An ordinary exclusion or cost omitted from the next mission is not inherited. Permanent keepouts and scheduled restrictions have their own lifetime and remain governed by $r$. A rejected update preserves the prior mission and produces no goal or policy projection for that candidate. Timers and event processing continue normally.

Commands received during a mission enter a first-in, first-out queue. The bridge checks a queued candidate again before committing it, because restrictions may have changed while it waited. A replacement is rechecked before the active goal is cancelled. Applying the same normalized static policy again preserves its content, although counters and repeated navigation requests can change. These semantics are examined by the implementation checks in Section~\ref{sec:boundary}.

\newpage
\subsection{Projecting the applicable policy into Nav2}
The bridge sends goals through \code{/follow\_waypoints} and supplies three global-costmap adapters. Each adapter uses the global map frame and costmap resolution. In the source code, KeepoutLayer and SoftCostLayer are named \code{KeepoutCommandLayer} and \code{ZoneSoftCostLayer}. Policy costs shape the global path; they are not supplied to the local costmap, so this configuration does not enforce policy restrictions as hard constraints on local-controller trajectories~\citep{nav2keepoutguide}.

\paragraph{KeepoutLayer.}
The bridge evaluates active zone restrictions and publishes the resulting list on \code{/forbidden\_zones\_update}. The layer assigns lethal-obstacle cost 254 to the corresponding grid cells. Expiry or an event permission removes that policy contribution. Sensor obstacles remain active. Native KeepoutFilter offers the equivalent basic keepout mechanism through occupancy-grid masks~\citep{nativekeepout}.

\paragraph{SoftCostLayer.}
Integer values on \code{/zone\_cost\_overrides} assign traversal costs from 0 to 252. Higher values discourage traversal without introducing a lethal region. Nav2 reserves 253 for inscribed obstacles, 254 for lethal obstacles, and 255 for unknown space. NavFn treats 253 as an obstacle, so it is excluded from the soft-cost range. Invalid direct cost messages are rejected while the preceding valid costs remain in effect.

\paragraph{ObjectAvoidanceLayer.}
YOLOv8 detections and RGB-D depth provide object locations transformed into the map frame. The layer receives them on \code{/object\_world\_positions} and applies cost 254 within the policy radius supplied through \code{/object\_avoidance\_radius}. Detection persistence and expiry govern removal of the contribution. The experiments use person detections. The configured radius sets the spatial extent of the policy cost region.

For known cells, the combination before inflation is
\begin{equation}
 C_{\mathrm{pre}}(x)=\max\{C_{\mathrm{sensor}}(x),C_{\mathrm{keepout}}(x),
 C_{\mathrm{soft}}(x),C_{\mathrm{object}}(x)\}.
 \label{eq:costs}
\end{equation}
The layer order is static, obstacle, keepout, soft cost, object avoidance, and inflation. Inflation preserves higher known costs, so its gradient can mask a smaller soft cost. Unknown space follows Nav2's \code{updateWithMax} behavior: an unknown layer value contributes nothing, and a known layer value can replace an unknown master value~\citep{nav2costlayer}. Equation~\eqref{eq:costs} applies only to known cells. A low soft cost cannot clear an obstacle.

\subsection{Time, events, and more than one robot}
A timed keepout remains separate from ordinary mission replacement. Wall-clock endpoints are mapped to the node's ROS clock and activated or expired through ROS timers. Simulation shares \code{/clock} with Nav2; the physical setup uses a common system clock. An event rule can permit passage through a normally forbidden zone. Its activation removes the applicable policy keepout while retaining sensor costs and restrictions configured to disallow overrides.

Shared policy-derived updates are distributed through ROS~2 topics to each robot's adapters. Every robot retains its own localization, sensor layers, planner, and controller. Matching policy regions can therefore coexist with different complete costmaps and paths. The two-robot experiments examine this common-policy projection.

\section{Interpreting Instructions and Containing Invalid Updates}
\label{sec:interpretation}
The interface evaluation measures conversion into the intended policy, including correct acceptance and accepted semantic errors. Fault injection tests rejection and preservation of the active mission.

\newpage
\subsection{Commands, models, and scoring}
The benchmark contains 150 English commands built from ten policy intentions, five phrasing levels, and three variants per intention--level combination. The intentions cover keepouts, ordered destinations, numeric zone costs, required transit, person avoidance with default or explicit radii, timed restrictions, event overrides, combined keepout/transit requirements, and multi-field policies. L0 is direct wording, L1 uses workplace phrasing, L2 redistributes or reorders clauses, L3 includes corrections or distracting content, and L4 expresses the requirement indirectly.

Four frozen local models are evaluated: Llama~3.1-8B, Qwen2.5-7B, Qwen3-8B, and Qwen3.5-9B. Each uses the same schema prompt and 15 input--output demonstrations, with temperature~0, \code{top\_p}=0, and seed~42. A deterministic keyword-and-slot parser provides the comparison. It extracts the same policy categories through fixed patterns. None of the demonstration sentences appears verbatim in the benchmark. The research team developed the prompt, demonstrations, parser, reference policies, and AI-assisted command set.

A correct policy must match the reference destinations and order, transit zones, exclusions and timing, costs, object radii, and event rules without adding an executable element. Correct acceptance compares the full normalized policy with the reference after admission by the validator. Raw accuracy evaluates the generated policy before validation, retaining additional or malformed entries (\supp, Section~\ref{supp:scoring}). Every command has one processing outcome: correct acceptance, incorrect acceptance, interpreter rejection, parsing rejection, or validator rejection. Explicit interpreter rejection is counted first and is not counted again at the validator.

Five fixed-setting calls per command produce unchanged aggregate counts. The analysis uses 150 commands as its denominator. Nominal Wilson intervals describe command-level proportions but do not model shared intentions, paraphrase dependence, or development bias. Latency measures candidate generation from submission on an AMD Ryzen~5 5600X workstation with 32~GB RAM and an RTX~3060 with 8~GB VRAM.

\subsection{Conversion accuracy and accepted semantic errors}
\begin{table}[htbp]
\centering\small
\caption{Policy conversion on 150 commands per method. Correct and incorrect columns both count accepted policies. The nominal 95\% Wilson interval applies to correct acceptance.}
\label{tab:conversion}
\begin{tabular}{@{}lrrrr@{}}\toprule
Method & Correct & Incorrect & 95\% CI (\%) & Mean latency \\\midrule
Rule-based parser & 44 & 101 & [22.6, 37.1] & 0.175 ms \\
Llama 3.1-8B & 107 & 20 & [63.6, 78.0] & 2.533 s \\
Qwen2.5-7B & 122 & 10 & [74.3, 86.8] & 1.694 s \\
Qwen3-8B & 141 & 7 & [89.0, 96.8] & 1.800 s \\
Qwen3.5-9B & 142 & 3 & [89.8, 97.3] & 3.011 s \\\bottomrule
\end{tabular}
\end{table}

\begin{figure}[htbp]
\centering\includegraphics[width=.95\linewidth]{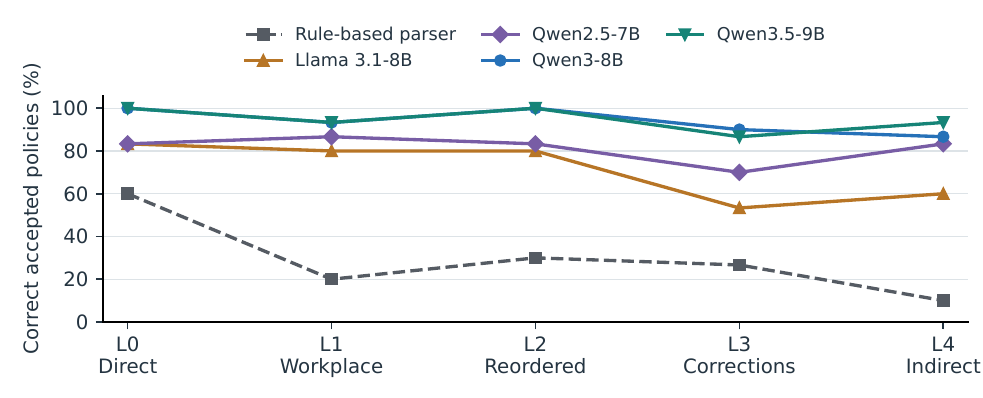}
\caption{Post-validation exact-match accuracy across five phrasing categories, with 30 commands per category. Exact counts and the complete acceptance/rejection breakdown are provided in the \supp.}
\label{fig:language}
\end{figure}

Correct acceptances range from 107/150 to 142/150 for the LLMs, compared with 44/150 for the parser (Table~\ref{tab:conversion}). The parser scores 60.0\% at L0 and 10.0--30.0\% at the other levels. Qwen3.5-9B and Qwen3-8B achieve similar observed accuracy, with 142 and 141 correct acceptances, and retain greater accuracy across phrasing variations (Figure~\ref{fig:language}).

Among accepted policies, 20/127 (15.7\%), 10/132 (7.6\%), 7/148 (4.7\%), and 3/145 (2.1\%) are incorrect for the four LLMs, respectively. Six of these 40 model--command cases omit a requested keepout or person-avoidance rule. Section~\ref{supp:error-types} classifies their potential operational effects.

LLM generation latency is 1.694--3.011~s, versus 0.175~ms for the parser. This cost occurs outside the planner and controller update cycles.

\newpage
\subsection{Fault containment and successive missions}
\label{sec:boundary}
The software fault-injection evaluation separates eight endpoint/JSON failures from 29 validator-stage faults. The latter comprise eight schema/type errors, eleven domain/range errors, and ten time/conflict errors. Containment requires rejection of the whole update, preservation of the preceding valid mission, and no new navigation goal or costmap projection for the rejected candidate.

\begin{table}[htbp]
\centering\small
\caption{Injected faults and their containment stage. All 37 faults preserve the prior mission state and produce no new policy projection. Six admissible controls are evaluated separately.}
\label{tab:faults}
\begin{tabular}{@{}lrrl@{}}\toprule
Fault class & Injected & Rejected & Stage \\\midrule
Endpoint/JSON & 8 & 8 & Before validation \\
Schema/type & 8 & 8 & Validator \\
Domain/range & 11 & 11 & Validator \\
Time/conflict & 10 & 10 & Validator \\\bottomrule
\end{tabular}
\end{table}

All 37 faults are contained, and six admissible controls produce the expected state behavior (Table~\ref{tab:faults}). A separate set of 100 contradictory instructions tests whether the interpreter recognizes a conflict before the structural validator sees it. Recognition ranges from 45/100 to 100/100 across the LLMs, and rejection at the policy-acceptance boundary ranges from 96/100 to 100/100. The parser scores 24/100 for both. If interpretation drops one conflicting clause, the validator receives an apparently consistent candidate. The category-level results are included in the \supp.

Software regression checks also cover mission replacement, retained keepouts, expiry, duplicate policy content, and replacement rejection before goal cancellation. These checks use instrumented interfaces; their configuration and results are detailed in the \supp.

\section{How Operating Policies Change Navigation}
\label{sec:navigation}
The navigation experiments use correct operating policies to isolate the route-selection effects of supplying operating constraints through the policy interface. The destination and the navigation algorithms are held fixed within each comparison. The experimental difference is the information supplied to the costmap, or, for the event task, whether a passage permission is active.

\newpage
\subsection{Testbeds and comparison conditions}
\begin{figure}[htbp]
\centering
\includegraphics[width=\linewidth]{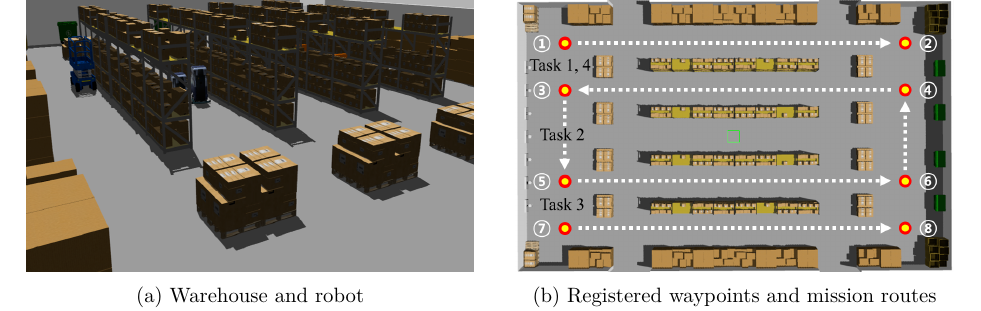}
\caption{Simulation testbed. Aisles are numbered from top to bottom. Tasks~1 and~4 connect WP1 to WP2; Task~2 follows WP5--WP6--WP4--WP3--WP5; Task~3 connects WP7 to WP8. Arrows indicate schematic mission segments.}
\label{fig:sim-env}
\end{figure}

The simulation uses a 25.9~m by 41.6~m warehouse in Gazebo. A DeliveryRobot model travels at up to 0.44~m/s with a 2D LiDAR (1-degree resolution, 5~Hz) and a forward-facing RGB-D camera. The parallel aisles provide alternative routes around shelves and task-specific obstructions (Figure~\ref{fig:sim-env}).

\begin{figure}[htbp]
\centering
\includegraphics[width=\linewidth]{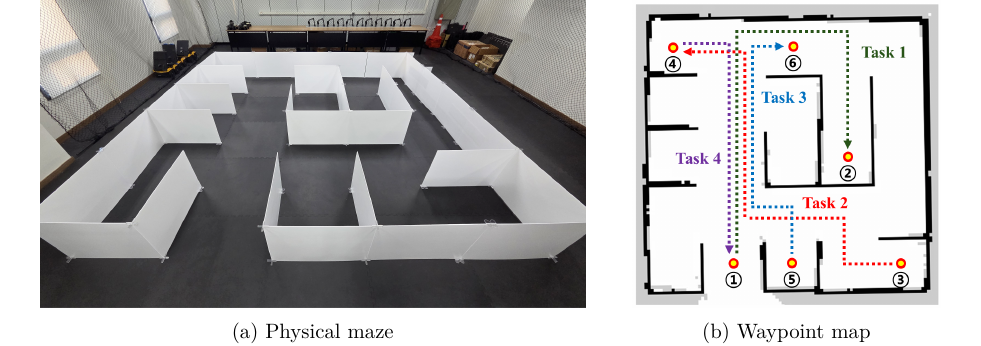}
\caption{Physical testbed. Task~1 connects WP1 to WP2, Task~2 WP3 to WP4, Task~3 WP5 to WP6, and Task~4 WP4 to WP1. The map shows the waypoint connections for the four tasks.}
\label{fig:real-env}
\end{figure}

The physical experiments use a TurtleBot3 Waffle in a maze, limited to 0.22~m/s, with an LDS-02 LiDAR, RealSense D435, and Jetson Orin Nano (Figure~\ref{fig:real-env}). YOLOv8 runs on the Jetson. Policy conversion and Nav2 run on the networked workstation.

Tasks~1--3 compare \emph{Baseline}, which uses NavFn, DWB, and standard sensor layers, with \emph{Policy}, which adds the accepted operating constraints through the adapters. Task~4 compares \emph{Priority-Off} and \emph{Priority-On} within the policy-enabled stack. Both retain the same sensor information; the latter activates a permission to cross a normally forbidden connection.

The navigation summaries report ten complete runs per condition. Start pose, destination, map, and task-specific obstacle conditions remain fixed. Time and traveled distance are measured from mission start to completion. Simulation Task~2 measures the entire multi-waypoint mission in both conditions.

Mean travel time and distance are compared using Welch's two-sample $t$-test, with 95\% confidence intervals for the mean differences (Tables~\ref{tab:nav-sim} and~\ref{tab:nav-physical}). Statistical details are provided in the \supp.

\newpage
\subsection{Avoiding a known closure and choosing among open aisles}
\begin{figure}[htbp]
\centering
\includegraphics[width=\linewidth]{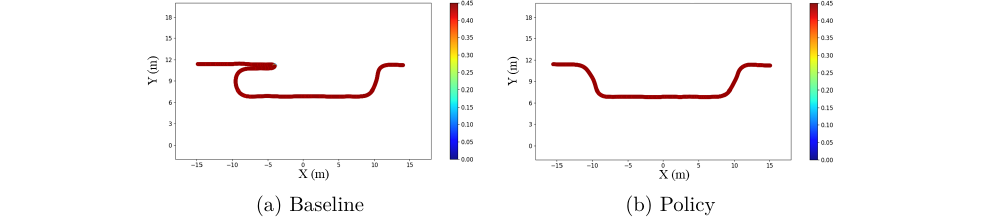}
\caption{Representative simulation Task~1 trajectories. Supplying the closure before navigation changes the initial route and avoids the approach to the blocked aisle.}
\label{fig:keepout-route}
\end{figure}

Task~1 places a blockage in the direct corridor. Baseline approaches the blocked route before taking an alternative. Declaring the region as a keepout allows the initial policy-informed path to bypass it, as illustrated by the simulation traces in Figure~\ref{fig:keepout-route}. Mean time falls from 114.009 to 86.125~s in simulation and from 62.179 to 40.390~s in the physical maze. The respective traveled distances decrease from 48.572 to 37.245~m and from 11.993 to 8.138~m.

Task~2 combines route choice with a longer mission. In simulation, exclusions close the third and fourth aisles and differential costs express preferences among the remaining alternatives. The complete mission changes from 311.369 to 238.751~s and from 126.152 to 96.003~m. In the physical task, zone costs favor the corridor containing a small box over the corridor occupied by another AMR. Mean time decreases from 55.849 to 42.959~s and distance from 10.850 to 8.303~m. The corresponding trajectory comparisons for both environments appear in the \supp.

\subsection{Applying a person-dependent cost region}
\begin{figure}[htbp]
\centering
\includegraphics[width=\linewidth]{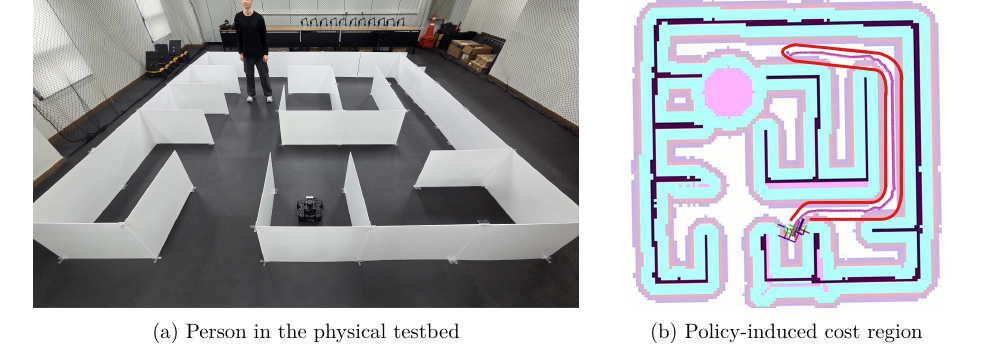}
\caption{Task~3 in the physical maze. A detected person activates the policy-defined object region in the costmap used for planning.}
\label{fig:person}
\end{figure}

In Task~3, a person detection activates the configured avoidance region. Figure~\ref{fig:person} places the physical scene beside the resulting RViz costmap, making the connection between detection and the planning representation visible. The additional region changes which route the robot selects.

Mean time changes from 142.049 to 115.872~s in simulation and from 42.109 to 32.160~s in the physical experiment. Distance decreases from 61.185 to 49.527~m and from 8.020 to 6.394~m, respectively. These are route-selection results for the tested person configuration.

\newpage
\subsection{Opening a passage when an event changes the policy}
\begin{figure}[htbp]
\centering
\includegraphics[width=\linewidth]{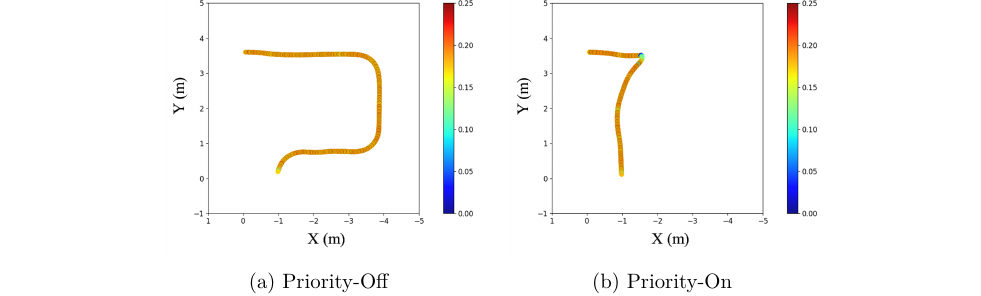}
\caption{Representative physical Task~4 trajectories. The active passage permission allows a more direct route within the same policy-enabled stack. Priority-Off and Priority-On differ only in this policy permission.}
\label{fig:priority-route}
\end{figure}

Task~4 tests a change in permission. A fire-alarm state allows a connection that the ordinary policy forbids. Sensor-derived obstacles remain active in both conditions. The physical Priority-Off route must follow a long detour, and Priority-On can use the released passage (Figure~\ref{fig:priority-route}). Time falls from 46.280 to 26.640~s and distance from 9.308 to 5.165~m. The simulation geometry produces a smaller change, from 86.170 to 81.056~s and from 37.241 to 34.699~m.

\begin{samepage}
Across Tasks~1--3, mean time reductions are 18.4--24.5\% in simulation and 23.1--35.0\% physically. The separate priority contrast yields 5.9\% and 42.4\%. The approximate 95\% intervals in Tables~\ref{tab:nav-sim} and~\ref{tab:nav-physical} exclude zero for all time and distance comparisons. The gains arise from the supplied operating information and its effect on Nav2 costs.

\end{samepage}

\begin{table}[htbp]
\centering\small
\caption{Simulation navigation: travel time and distance (mean $\pm$ sample SD, $n=10$ per condition). Intervals are approximate 95\% Welch confidence intervals for the indicated difference, in the metric's units.}
\label{tab:nav-sim}
\begin{tabular}{@{}clrrr@{}}\toprule
Task & Metric & Baseline & Policy & 95\% CI: Baseline $-$ Policy \\\midrule
1 & Time (s) & $114.009 \pm 0.721$ & $86.125 \pm 0.230$ & [27.356, 28.412] \\
1 & Distance (m) & $48.572 \pm 0.108$ & $37.245 \pm 0.108$ & [11.226, 11.428] \\
2 & Time (s) & $311.369 \pm 1.444$ & $238.751 \pm 3.643$ & [69.912, 75.324] \\
2 & Distance (m) & $126.152 \pm 0.180$ & $96.003 \pm 0.177$ & [29.981, 30.317] \\
3 & Time (s) & $142.049 \pm 0.315$ & $115.872 \pm 0.790$ & [25.590, 26.764] \\
3 & Distance (m) & $61.185 \pm 0.140$ & $49.527 \pm 0.340$ & [11.405, 11.911] \\
\midrule
Task & Metric & Priority-Off & Priority-On & 95\% CI: Off $-$ On \\\midrule
4 & Time (s) & $86.170 \pm 0.456$ & $81.056 \pm 0.328$ & [4.738, 5.490] \\
4 & Distance (m) & $37.241 \pm 0.186$ & $34.699 \pm 0.121$ & [2.393, 2.691] \\
\bottomrule\end{tabular}
\end{table}

\begin{table}[htbp]
\centering\small
\caption{Physical-testbed navigation: travel time and distance (mean $\pm$ sample SD, $n=10$ per condition). Intervals are approximate 95\% Welch confidence intervals for the indicated difference, in the metric's units.}
\label{tab:nav-physical}
\begin{tabular}{@{}clrrr@{}}\toprule
Task & Metric & Baseline & Policy & 95\% CI: Baseline $-$ Policy \\\midrule
1 & Time (s) & $62.179 \pm 1.322$ & $40.390 \pm 1.213$ & [20.596, 22.982] \\
1 & Distance (m) & $11.993 \pm 0.210$ & $8.138 \pm 0.190$ & [3.667, 4.043] \\
2 & Time (s) & $55.849 \pm 0.687$ & $42.959 \pm 1.079$ & [12.029, 13.751] \\
2 & Distance (m) & $10.850 \pm 0.080$ & $8.303 \pm 0.156$ & [2.428, 2.666] \\
3 & Time (s) & $42.109 \pm 2.073$ & $32.160 \pm 1.153$ & [8.341, 11.557] \\
3 & Distance (m) & $8.020 \pm 0.352$ & $6.394 \pm 0.216$ & [1.348, 1.904] \\
\midrule
Task & Metric & Priority-Off & Priority-On & 95\% CI: Off $-$ On \\\midrule
4 & Time (s) & $46.280 \pm 0.341$ & $26.640 \pm 0.230$ & [19.364, 19.916] \\
4 & Distance (m) & $9.308 \pm 0.047$ & $5.165 \pm 0.018$ & [4.108, 4.178] \\
\bottomrule\end{tabular}
\end{table}

\section{Checking the Nav2 Integration}
\label{sec:integration}
The remaining experiments isolate the basic keepout function and examine two behaviors that extend beyond a single static mission: expiration of a restriction and distribution of a policy to two robots.

\subsection{Native KeepoutFilter and the custom adapter}
A direct control bypasses the LLM, validator, and bridge. Native KeepoutFilter receives an occupancy-grid mask and \code{CostmapFilterInfo}; custom KeepoutLayer receives the equivalent Zone~A restriction on \code{/forbidden\_zones\_update}. The comparison uses ROS~2 Humble, Nav2~1.1.20, Gazebo Classic~11, and the \code{factory6} map, with WP2 as the destination.

Five runs per condition independently restart Gazebo, the robot, AMCL, and Nav2. Conditions alternate and the simulation seed is fixed at 42. Mask raster boundaries, global resolution (0.05~m), inflation radius (1.3~m), scaling factor (3.0), and robot radius (0.65~m) are matched. Native applies an additional InflationLayer after KeepoutFilter in the filter sequence; Custom applies inflation after its keepout plugin. The \supp\ gives the complete measurement definitions and pre-motion procedure.

\begin{table}[htbp]
\centering\small
\caption{Direct keepout control, with five restarted runs per condition. Values are mean $\pm$ sample SD. Travel time starts at first detected motion and ends at the waypoint-action completion result.}
\label{tab:native}
\begin{tabular}{@{}lcrr@{}}\toprule
Method & Completed & Distance (m) & Time (s) \\\midrule
Native KeepoutFilter & 5/5 & $37.339\pm0.103$ & $86.430\pm0.236$ \\
Custom KeepoutLayer & 5/5 & $37.300\pm0.131$ & $86.351\pm0.329$ \\\bottomrule
\end{tabular}
\end{table}

Both methods complete every run (Table~\ref{tab:native}). In pre-motion planning checks, both methods produce paths that avoid the forbidden zone, cross it after release, and avoid it again after reapplication. Five costmap comparisons record no differences among 45,384 sampled cells in the zone and its 1.3~m surrounding region. Release restores those sampled costs, including the inflation region. Both implementations applied the tested keepout updates successfully and produced similar mean travel distances and times.

\subsection{Restriction expiry and shared policies}
A simulation sequence follows a timed keepout through activation, expiry, and replanning. The robot initially detours around Zone~A. After the timer expires, the same destination remains active and Nav2 replans through the released region. The three costmap views are included in the \supp.

\begin{figure}[htbp]
\centering
\includegraphics[width=\linewidth]{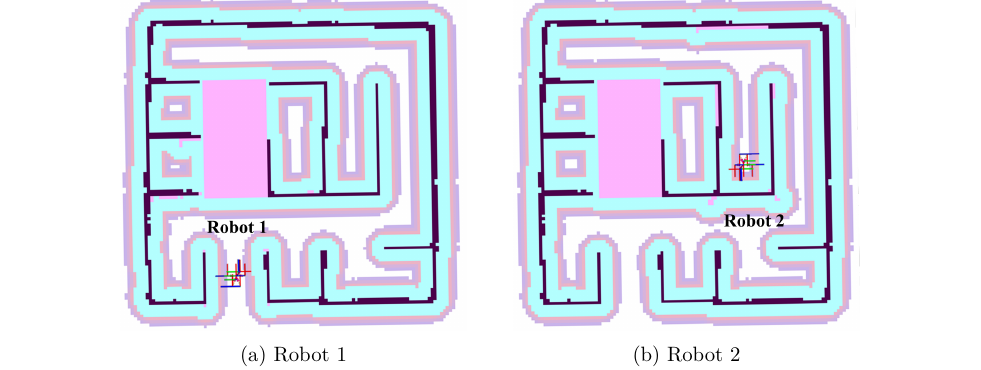}
\caption{Shared policy projection in the physical two-robot demonstration. Matching policy-induced regions are visible in both costmaps. Each robot retains its own sensor layers, localization, planner, and controller.}
\label{fig:sharing}
\end{figure}

Shared-policy demonstrations use two DeliveryRobots in simulation and two TurtleBot3 robots in the physical testbed. Each pair receives common policy-derived updates and runs independent navigation instances. Figure~\ref{fig:sharing} shows the physical pair's matching policy regions. The simulated pair is documented in the \supp. Local sensor costs and paths can differ because each robot plans from its own state.

\section{Discussion}
An operating rule has a lifetime beyond the sentence that introduced it. The interface makes that lifetime explicit through mission replacement, retained restrictions, timers, and event state. This is particularly useful when the next instruction arrives before a previous restriction expires. The context check can reject a required transit zone that is currently closed without silently changing the requested mission.

The observed missing keepouts, omitted stops, and added goals motivate previews that place the instruction beside the proposed policy. Confirmation of critical restriction changes, targeted clarification of numeric roles, and authorization checks for persistent restrictions are candidate extensions for operator review. The benchmark, prompt, parser, and reference policies share a development process; evaluation with independently authored commands is needed to assess generalization.

The navigation experiments examine operating policies in two fixed layouts. The baseline lacks operating information that the policy-enabled condition receives. The navigation gains reflect the incorporation of this information into Nav2's costmap. Small run-to-run SDs describe these conditions. Additional layouts, obstacle behaviors, and instruction sources are needed to assess generalization. The configured radius specifies the costmap region; achieved clearance depends on perception and navigation execution.

\clearpage
\section{Conclusion}
We developed a policy interface that connects operator instructions to Nav2 through structured policies, deterministic acceptance, and explicit mission-state handling. The language benchmark quantifies both successful conversions and accepted semantic errors. Simulation and physical experiments demonstrate route changes under operating policies. In the direct keepout comparison, both implementations applied the tested prohibition, release, and reapplication updates successfully. Further evaluation should jointly measure mission completion and instruction compliance from natural-language input through robot execution.

\noindent\textbf{Data and code.} Code, benchmark data, and demonstrations are accessible through the \href{\ProjectURL}{project page}. Implementation versions and additional experimental details are provided in the \supp.

\ifanonymous\else\relax
\fi
\FloatBarrier
\begingroup\footnotesize
\setlength{\bibsep}{0pt plus 1pt}
\bibliographystyle{unsrtnat}
\bibliography{references}

\begin{thebibliography}{15}
\providecommand{\natexlab}[1]{#1}
\providecommand{\url}[1]{\texttt{#1}}
\expandafter\ifx\csname urlstyle\endcsname\relax
  \providecommand{\doi}[1]{doi: #1}\else
  \providecommand{\doi}{doi: \begingroup \urlstyle{rm}\Url}\fi

\bibitem[da~Costa~Barros and Nascimento(2021)]{barros2021fulfillment}
{\'I}talo~Renan da~Costa~Barros and Tiago~Pereira Nascimento.
\newblock Robotic mobile fulfillment systems: A survey on recent developments
  and research opportunities.
\newblock \emph{Robotics and Autonomous Systems}, 137:\penalty0 103729, 2021.
\newblock \doi{10.1016/j.robot.2021.103729}.

\bibitem[Cen~Cheng et~al.(2022)Cen~Cheng, Indri, Sibona, De~Rose, and
  Prato]{cheng2022dynamic}
Pangcheng~David Cen~Cheng, Marina Indri, Fiorella Sibona, Matteo De~Rose, and
  Gianluca Prato.
\newblock Dynamic path planning of a mobile robot adopting a costmap layer
  approach in {ROS2}.
\newblock In \emph{IEEE International Conference on Emerging Technologies and
  Factory Automation (ETFA)}, pages 1--8, 2022.
\newblock \doi{10.1109/ETFA52439.2022.9921458}.
\newblock URL \url{https://iris.polito.it/handle/11583/2970835}.

\bibitem[Pakkala et~al.(2025)Pakkala, K{\"a}ns{\"a}koski, Heikkil{\"a},
  Backman, and P{\"a}{\"a}kk{\"o}nen]{pakkala2025design}
Daniel Pakkala, Niko K{\"a}ns{\"a}koski, Tapio Heikkil{\"a}, Jere Backman, and
  Pekka P{\"a}{\"a}kk{\"o}nen.
\newblock On design of cognitive situation-adaptive autonomous mobile robotic
  applications.
\newblock \emph{Computers in Industry}, 167:\penalty0 104263, 2025.
\newblock \doi{10.1016/j.compind.2025.104263}.

\bibitem[Macenski et~al.(2020)Macenski, Mart{\'i}n, White, and
  Clavero]{macenski2020marathon}
Steve Macenski, Francisco Mart{\'i}n, Ruffin White, and Jonatan~Gin{\'e}s
  Clavero.
\newblock The marathon 2: A navigation system.
\newblock In \emph{IEEE/RSJ International Conference on Intelligent Robots and
  Systems (IROS)}, pages 2718--2725, 2020.
\newblock \doi{10.1109/IROS45743.2020.9341207}.
\newblock URL \url{https://arxiv.org/abs/2003.00368}.

\bibitem[Lu et~al.(2014)Lu, Hershberger, and Smart]{lu2014layered}
David~V. Lu, Dave Hershberger, and William~D. Smart.
\newblock Layered costmaps for context-sensitive navigation.
\newblock In \emph{IEEE/RSJ International Conference on Intelligent Robots and
  Systems (IROS)}, pages 709--715, 2014.
\newblock \doi{10.1109/IROS.2014.6942636}.

\bibitem[{Nav2 Developers}(2026{\natexlab{a}})]{nativekeepout}
{Nav2 Developers}.
\newblock {Nav2} 1.1.20: {KeepoutFilter} implementation.
\newblock Versioned source code, 2026{\natexlab{a}}.
\newblock URL
  \url{https://github.com/ros-navigation/navigation2/blob/1.1.20/nav2_costmap_2d/plugins/costmap_filters/keepout_filter.cpp}.
\newblock Accessed 22 October 2026.

\bibitem[Koubaa(2023)]{koubaa2023rosgpt}
Anis Koubaa.
\newblock {ROSGPT}: Next-generation human-robot interaction with {ChatGPT} and
  {ROS}.
\newblock Preprints.org, version 2, 2023.
\newblock URL \url{https://www.preprints.org/manuscript/202304.0827/v2}.

\bibitem[Liang et~al.(2022)Liang, Huang, Xia, Xu, Hausman, Ichter, Florence,
  and Zeng]{liang2023code}
Jacky Liang, Wenlong Huang, Fei Xia, Peng Xu, Karol Hausman, Brian Ichter, Pete
  Florence, and Andy Zeng.
\newblock Code as policies: Language model programs for embodied control.
\newblock \emph{arXiv preprint arXiv:2209.07753}, 2022.
\newblock URL \url{https://arxiv.org/abs/2209.07753}.

\bibitem[Meng et~al.(2024)Meng, Wang, Yang, Peng, and Chang]{meng2024llm}
Silin Meng, Yiwei Wang, Cheng-Fu Yang, Nanyun Peng, and Kai-Wei Chang.
\newblock {LLM-A*}: Large language model enhanced incremental heuristic search
  on path planning.
\newblock In \emph{Findings of the Association for Computational Linguistics:
  EMNLP 2024}, pages 1087--1102, 2024.
\newblock \doi{10.18653/v1/2024.findings-emnlp.60}.
\newblock URL \url{https://aclanthology.org/2024.findings-emnlp.60/}.

\bibitem[Latif(2024)]{latif20243p}
Ehsan Latif.
\newblock {3P-LLM}: Probabilistic path planning using large language model for
  autonomous robot navigation.
\newblock \emph{arXiv preprint arXiv:2403.18778}, 2024.
\newblock URL \url{https://arxiv.org/abs/2403.18778}.

\bibitem[Chen et~al.(2022)Chen, Lin, Lai, and Lin]{chen2022velocity}
Chin~S. Chen, Chia~J. Lin, Chun~C. Lai, and Si~Y. Lin.
\newblock Velocity estimation and cost map generation for dynamic obstacle
  avoidance of {ROS} based {AMR}.
\newblock \emph{Machines}, 10\penalty0 (7):\penalty0 501, 2022.
\newblock URL \url{https://www.mdpi.com/2075-1702/10/7/501}.

\bibitem[Sani et~al.(2024)Sani, Sgorbissa, and Carpin]{sani2024improving}
Ettore Sani, Antonio Sgorbissa, and Stefano Carpin.
\newblock Improving the {ROS 2} navigation stack with real-time local costmap
  updates for agricultural applications.
\newblock In \emph{IEEE International Conference on Robotics and Automation
  (ICRA)}, pages 17701--17707, 2024.
\newblock URL \url{https://arxiv.org/abs/2407.18535}.

\bibitem[Zaj{\k{a}}c and Ma{\l}opolski(2021)]{zajac2021structural}
Jerzy Zaj{\k{a}}c and Waldemar Ma{\l}opolski.
\newblock Structural on-line control policy for collision and deadlock
  resolution in multi-{AGV} systems.
\newblock \emph{Journal of Manufacturing Systems}, 60:\penalty0 80--92, 2021.
\newblock \doi{10.1016/j.jmsy.2021.05.002}.

\bibitem[{Nav2 Developers}(2026{\natexlab{b}})]{nav2keepoutguide}
{Nav2 Developers}.
\newblock Navigating with keepout zones.
\newblock Official Nav2 documentation, 2026{\natexlab{b}}.
\newblock URL
  \url{https://docs.nav2.org/rolling/tutorials/general_tutorials/navigation2_with_keepout_filter/navigation2_with_keepout_filter/}.
\newblock Section 3: Enable Keepout Filter. Accessed 23 October 2026.

\bibitem[{Nav2 Developers}(2026{\natexlab{c}})]{nav2costlayer}
{Nav2 Developers}.
\newblock {Nav2} 1.1.20: {CostmapLayer} implementation.
\newblock Versioned source code, 2026{\natexlab{c}}.
\newblock URL
  \url{https://github.com/ros-navigation/navigation2/blob/1.1.20/nav2_costmap_2d/src/costmap_layer.cpp}.
\newblock Accessed 22 October 2026.

\end{thebibliography}
\endgroup
\clearpage
\appendix
\renewcommand{\thesection}{S\arabic{section}}
\setcounter{section}{0}
\setcounter{figure}{0}
\setcounter{table}{0}
\setcounter{equation}{0}
\renewcommand{\thefigure}{S\arabic{figure}}
\renewcommand{\theHfigure}{supp.\arabic{figure}}
\renewcommand{\theHtable}{supp.\arabic{table}}
\renewcommand{\theHequation}{supp.\arabic{equation}}
\renewcommand{\thetable}{S\arabic{table}}
\renewcommand{\theequation}{S\arabic{equation}}
\section*{Supplementary Material}
\label{supp:start}
This supplement provides detailed language outcomes, implementation checks, statistical calculations, and additional experimental views for \emph{Natural-Language to Nav2: Operational Policies for Costmap-Based Navigation}.

\section{Language benchmark details}
\label{supp:language}
Table~\ref{tab:language} includes raw policy matches, correct acceptances, nominal Wilson intervals, and generation latency. Table~\ref{tab:levels} supplies the proportions used in Figure~\ref{fig:language}; each level has 30 commands. The outcome partition in Table~\ref{tab:outcomes} distinguishes incorrect acceptance from rejection. All rows sum to 150.
\begin{table}[htbp]
\centering\small
\caption{Policy-conversion results on 150 commands. Raw and Post-V count policies matching the reference before and after validation, respectively. The 95\% Wilson interval is nominal and applies to Post-V. Latency concerns candidate generation only.}
\label{tab:language}
\begin{tabular}{@{}lrrrrl@{}}\toprule
Method & Raw & Post-V & Accuracy (\%) & 95\% CI (\%) & Latency \\\midrule
Rule-based parser & 44/150 & 44/150 & 29.3 & [22.6, 37.1] & 0.175 ms \\
Llama 3.1-8B & 106/150 & 107/150 & 71.3 & [63.6, 78.0] & 2.533 s \\
Qwen2.5-7B & 122/150 & 122/150 & 81.3 & [74.3, 86.8] & 1.694 s \\
Qwen3-8B & 141/150 & 141/150 & 94.0 & [89.0, 96.8] & 1.800 s \\
Qwen3.5-9B & 142/150 & 142/150 & 94.7 & [89.8, 97.3] & 3.011 s \\
\bottomrule\end{tabular}
\end{table}
\begin{table}[htbp]
\centering\small
\caption{Post-validation exact-match accuracy (\%) by phrasing level, with 30 commands per level.}
\label{tab:levels}
\begin{tabular}{@{}lrrrrr@{}}\toprule
Method & L0 & L1 & L2 & L3 & L4 \\\midrule
Rule-based parser & 60.0 & 20.0 & 30.0 & 26.7 & 10.0 \\
Llama 3.1-8B & 83.3 & 80.0 & 80.0 & 53.3 & 60.0 \\
Qwen2.5-7B & 83.3 & 86.7 & 83.3 & 70.0 & 83.3 \\
Qwen3-8B & 100.0 & 93.3 & 100.0 & 90.0 & 86.7 \\
Qwen3.5-9B & 100.0 & 93.3 & 100.0 & 86.7 & 93.3 \\
\bottomrule\end{tabular}
\end{table}

\begin{table}[htbp]
\centering\small
\caption{Mutually exclusive processing outcomes, each row summing to 150. Interpreter rejection denotes an explicit model or parser rejection.}
\label{tab:outcomes}
\begin{tabular}{@{}lrrrrr@{}}\toprule
& \multicolumn{2}{c}{Accepted} & \multicolumn{3}{c}{Rejected by} \\
\cmidrule(lr){2-3}\cmidrule(l){4-6}
Method & Correct & Incorrect & Interpreter & Parsing & Validator \\\midrule
Rule-based parser & 44 & 101 & 5 & 0 & 0 \\
Llama 3.1-8B & 107 & 20 & 0 & 0 & 23 \\
Qwen2.5-7B & 122 & 10 & 3 & 0 & 15 \\
Qwen3-8B & 141 & 7 & 1 & 0 & 1 \\
Qwen3.5-9B & 142 & 3 & 4 & 0 & 1 \\
\bottomrule\end{tabular}
\end{table}

\subsection{Contradictory instructions}
Each contradiction category contains 20 commands. C1 combines forbidden and required transit through a zone; C2 specifies simultaneous open/closed requirements; C3 requests incompatible sole final destinations; C4 contains conflicting rules for the same event; C5 has overlapping timed conflicts. Each command is evaluated five times at fixed settings; intervals use the 100 distinct commands as the denominator.

\begin{table}[htbp]
\centering\small
\caption{Contradiction recognition by category (\%, 20 commands each) and overall counts out of 100. Recognized denotes explicit contradiction recognition; Blocked denotes rejection at the policy-acceptance boundary.}
\label{tab:contradiction}
\begin{tabular}{@{}lrrrrrrr@{}}\toprule
Method & C1 & C2 & C3 & C4 & C5 & Recognized & Blocked \\\midrule
Rule-based parser & 20 & 10 & 0 & 40 & 50 & 24/100 & 24/100 \\
Llama 3.1-8B & 55 & 55 & 90 & 15 & 10 & 45/100 & 96/100 \\
Qwen2.5-7B & 85 & 90 & 90 & 55 & 65 & 77/100 & 97/100 \\
Qwen3-8B & 100 & 90 & 100 & 85 & 85 & 92/100 & 99/100 \\
Qwen3.5-9B & 100 & 100 & 100 & 100 & 100 & 100/100 & 100/100 \\
\bottomrule\end{tabular}
\end{table}
\begin{table}[htbp]
\centering\small
\caption{Nominal 95\% Wilson intervals (\%) for the 100-command contradiction results. The same command-level and shared-category limitations apply as in the conversion evaluation.}
\label{tab:contradiction-ci}
\begin{tabular}{@{}lrr@{}}\toprule
Method & Recognition & Boundary containment \\\midrule
Rule-based parser & [16.7, 33.2] & [16.7, 33.2] \\
Llama 3.1-8B & [35.6, 54.8] & [90.2, 98.4] \\
Qwen2.5-7B & [67.8, 84.2] & [91.5, 99.0] \\
Qwen3-8B & [85.0, 95.9] & [94.6, 99.8] \\
Qwen3.5-9B & [96.3, 100.0] & [96.3, 100.0] \\
\bottomrule\end{tabular}
\end{table}

Recognition requires \code{decision=reject} and \code{reason\_code=contradiction}. Boundary containment is assessed offline from recorded rejection and unchanged policy-state and version flags.

Qwen3.5-9B recognizes and blocks all 100 commands; 72 responses also satisfy the rejection format requiring all six policy arrays to be empty.

An offline check of the released parser and validator reproduced 24 explicit rejections and 76 accepted candidates on the contradiction set. On the 150-command conversion set, they reproduce 44 correct acceptances, 101 incorrect acceptances, and five explicit rejections. When both exclusion and transit are extracted for the same zone, the parser can reject the conflict itself. In the examined accepted C1 and C5 failures, at least one conflicting requirement is absent from the candidate, leaving no corresponding structural conflict for the validator to reject.

\FloatBarrier
\section{Mission-state and timing checks}
\label{supp:checks}
The Python suite passes 102 tests on the tested implementation. The eight node-level cases in Table~\ref{tab:transitions} check validation, queue handling, commit, goal construction, and timers using instrumented substitutes for the interpreter, publishers, and navigation interfaces. These node-level tests verify state transitions and outgoing software requests.

\begin{table}[htbp]
\centering\small
\caption{Node-level checks on the tested implementation. All eight expected outcomes are observed.}
\label{tab:transitions}
\begin{tabularx}{\linewidth}{@{}lp{49mm}X@{}}\toprule
Case & Input and context & Expected behavior \\\midrule
T1 & Valid policy & Accept constraints and construct the requested goal sequence. \\
T2 & New mission & Replace the snapshot without inheriting omitted ordinary constraints. \\
T3a & Exclusion and required transit in one policy & Reject the update and preserve the preceding mission. \\
T3b & Required transit under a timed keepout & Reject without silently deleting required transit. \\
T3c & Required transit under a permanent keepout & Reject and retain the preceding mission. \\
T3d & Required transit after expiry & Accept the candidate with transit retained. \\
T4 & Out-of-range zone cost & Reject the entire update. \\
T5 & Identical static policy applied twice & Preserve normalized content; counters may still change. \\\bottomrule
\end{tabularx}
\end{table}

A one-second ROS timer checks expiry. The 43 fault/control checks comprise eight endpoint/JSON faults, 29 validator faults, three normalization controls, one state-content control, and two semantic-boundary controls that accept structurally valid policies with incorrect intent. These checks overlap the unit suite.

\begin{figure}[htbp]
\centering
\includegraphics[width=\linewidth]{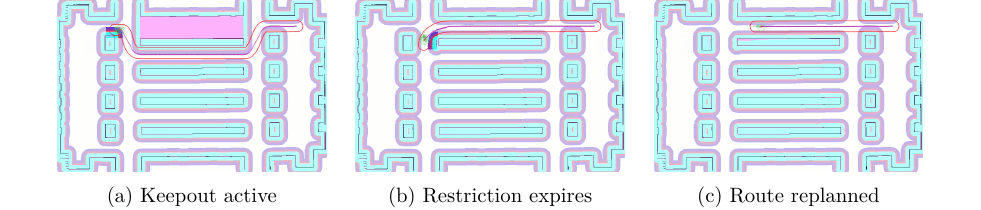}
\caption{Expiry of a Zone~A keepout during one simulation run. The destination remains active as the policy contribution is removed and Nav2 selects a route through the released region.}
\label{fig:expiry}
\end{figure}

\FloatBarrier
\section{Statistical calculations and the direct keepout control}
\label{supp:statistics}
Tables~\ref{tab:simulation} and~\ref{tab:physical} provide the full mission results. 

The reported sample size is ten runs per condition ($n_1=n_2=10$). Let $\Delta=\bar{x}_1-\bar{x}_2$, $a=s_1^2/n_1$, and $b=s_2^2/n_2$. The Welch test statistic, degrees of freedom, and 95\% confidence interval are
\begin{equation}
 t=\frac{\Delta}{\sqrt{a+b}},\qquad
 \nu=\frac{(a+b)^2}{a^2/(n_1-1)+b^2/(n_2-1)},\qquad
 \mathrm{CI}_{95\%}=\Delta\pm t_{0.975,\nu}\sqrt{a+b}.
\end{equation}
The two-sided $p$-value is $2\Pr(T_\nu\geq|t|)$. Table~\ref{tab:welch} reports the test results. No multiple-comparison adjustment is applied to $p$-values or confidence intervals.
\begin{table}[htbp]
\centering\small
\caption{Simulation navigation: mean $\pm$ sample SD, $n=10$ per condition. Approximate 95\% Welch intervals are for the difference between conditions.}
\label{tab:simulation}
\begin{tabular}{@{}clrrr@{}}\toprule
Task & Metric & Baseline & Policy & 95\% CI: Baseline $-$ Policy \\\midrule
1 & Time (s) & $114.009 \pm 0.721$ & $86.125 \pm 0.230$ & [27.356, 28.412] \\
1 & Distance (m) & $48.572 \pm 0.108$ & $37.245 \pm 0.108$ & [11.226, 11.428] \\
2 & Time (s) & $311.369 \pm 1.444$ & $238.751 \pm 3.643$ & [69.912, 75.324] \\
2 & Distance (m) & $126.152 \pm 0.180$ & $96.003 \pm 0.177$ & [29.981, 30.317] \\
3 & Time (s) & $142.049 \pm 0.315$ & $115.872 \pm 0.790$ & [25.590, 26.764] \\
3 & Distance (m) & $61.185 \pm 0.140$ & $49.527 \pm 0.340$ & [11.405, 11.911] \\
\midrule
Task & Metric & Priority-Off & Priority-On & 95\% CI: Off $-$ On \\\midrule
4 & Time (s) & $86.170 \pm 0.456$ & $81.056 \pm 0.328$ & [4.738, 5.490] \\
4 & Distance (m) & $37.241 \pm 0.186$ & $34.699 \pm 0.121$ & [2.393, 2.691] \\
\bottomrule\end{tabular}
\end{table}

\begin{table}[htbp]
\centering\small
\caption{Physical navigation: mean $\pm$ sample SD, $n=10$ per condition. Approximate 95\% Welch intervals are for the difference between conditions.}
\label{tab:physical}
\begin{tabular}{@{}clrrr@{}}\toprule
Task & Metric & Baseline & Policy & 95\% CI: Baseline $-$ Policy \\\midrule
1 & Time (s) & $62.179 \pm 1.322$ & $40.390 \pm 1.213$ & [20.596, 22.982] \\
1 & Distance (m) & $11.993 \pm 0.210$ & $8.138 \pm 0.190$ & [3.667, 4.043] \\
2 & Time (s) & $55.849 \pm 0.687$ & $42.959 \pm 1.079$ & [12.029, 13.751] \\
2 & Distance (m) & $10.850 \pm 0.080$ & $8.303 \pm 0.156$ & [2.428, 2.666] \\
3 & Time (s) & $42.109 \pm 2.073$ & $32.160 \pm 1.153$ & [8.341, 11.557] \\
3 & Distance (m) & $8.020 \pm 0.352$ & $6.394 \pm 0.216$ & [1.348, 1.904] \\
\midrule
Task & Metric & Priority-Off & Priority-On & 95\% CI: Off $-$ On \\\midrule
4 & Time (s) & $46.280 \pm 0.341$ & $26.640 \pm 0.230$ & [19.364, 19.916] \\
4 & Distance (m) & $9.308 \pm 0.047$ & $5.165 \pm 0.018$ & [4.108, 4.178] \\
\bottomrule\end{tabular}
\end{table}

\begin{table}[htbp]
\centering\small
\caption{Approximate two-sided Welch test statistics. Differences are Baseline $-$ Policy for Tasks 1--3 and Priority-Off $-$ Priority-On for Task 4. No multiple-comparison adjustment is applied.}
\label{tab:welch}
\begin{tabular}{@{}clrrrr@{}}\toprule
Task & Metric & Difference & $t$ & Degrees of freedom & $p$ \\\midrule
\multicolumn{6}{c}{\textit{Simulation}} \\\midrule
1 & Time (s) & 27.884 & 116.5134 & 10.8129 & $4.165 \times 10^{-18}$ \\
1 & Distance (m) & 11.327 & 234.5180 & 18.0000 & $7.968 \times 10^{-33}$ \\
2 & Time (s) & 72.618 & 58.5999 & 11.7599 & $7.129 \times 10^{-16}$ \\
2 & Distance (m) & 30.149 & 377.6629 & 17.9949 & $1.535 \times 10^{-36}$ \\
3 & Time (s) & 26.177 & 97.3314 & 11.7912 & $1.689 \times 10^{-18}$ \\
3 & Distance (m) & 11.658 & 100.2618 & 11.9666 & $7.145 \times 10^{-19}$ \\
4 & Time (s) & 5.114 & 28.7904 & 16.3464 & $1.937 \times 10^{-15}$ \\
4 & Distance (m) & 2.542 & 36.2268 & 15.4605 & $2.253 \times 10^{-16}$ \\
\midrule
\multicolumn{6}{c}{\textit{Physical}} \\\midrule
1 & Time (s) & 21.789 & 38.4037 & 17.8683 & $1.264 \times 10^{-18}$ \\
1 & Distance (m) & 3.855 & 43.0464 & 17.8227 & $1.828 \times 10^{-19}$ \\
2 & Time (s) & 12.890 & 31.8664 & 15.2671 & $2.193 \times 10^{-15}$ \\
2 & Distance (m) & 2.547 & 45.9415 & 13.4275 & $3.684 \times 10^{-16}$ \\
3 & Time (s) & 9.949 & 13.2633 & 14.0821 & $2.385 \times 10^{-9}$ \\
3 & Distance (m) & 1.626 & 12.4504 & 14.9362 & $2.743 \times 10^{-9}$ \\
4 & Time (s) & 19.640 & 150.9961 & 15.7846 & $2.272 \times 10^{-26}$ \\
4 & Distance (m) & 4.143 & 260.3139 & 11.5845 & $3.469 \times 10^{-23}$ \\
\midrule
\end{tabular}
\end{table}

\subsection{Native/custom control protocol}
The \code{factory6} control uses start $(-0.20,0.41)$~m and goal $(32.20,0.41)$~m. Native and Custom share raster boundaries and the parameter values described in Section~\ref{sec:integration}. Native retains plugin-stage inflation and additionally applies InflationLayer after KeepoutFilter in the filter sequence. Custom applies inflation after KeepoutLayer in its plugin sequence. Other custom adapters receive no soft-cost or object-policy input.

\begin{samepage}
Each restarted run checks four states before motion: unmodified, forbidden, released, and re-forbidden. Each update is published three times and allowed to stabilize. The final mission uses the keepout. Time is measured in ROS simulation time from the first motion after goal acceptance ($|v|>0.01$~m/s) to the FollowWaypoints completion result, including stops. Distance is the sum of planar odometry displacement over that interval. Setup and pre-motion checks are excluded. This definition differs from mission-start timing in Tasks~1--4.\par
\end{samepage}

The comparison covers 45,384 cells in the restricted zone and a surrounding rectangle extending 1.3~m beyond it. Five snapshot comparisons record zero native/custom differences. These measurements cover the sampled region and pre-motion updates in one simulation case.

\clearpage
\section{Additional trajectory comparisons}
\label{supp:trajectories}
The six trajectory comparisons below complement simulation Task~1 and physical Task~4 in the main paper. Each panel shows a representative individual run. Tables~\ref{tab:simulation} and~\ref{tab:physical} summarize the conditions.

\subsection{Declared closure in the physical maze}
Task~1 connects WP1 to WP2. The policy closes the blocked corridor before route selection. The physical comparison has the same interpretation as the simulation example in the main paper.
\begin{figure}[htbp]
\centering
\includegraphics[width=\linewidth]{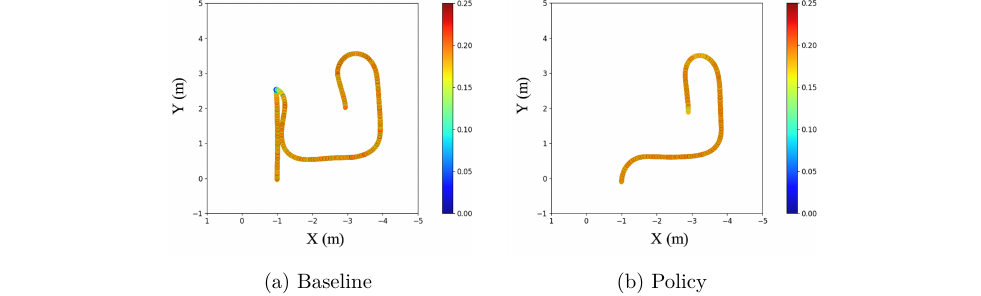}
\caption{Physical Task~1. The policy-informed route avoids the declared closure.}
\end{figure}

\subsection{Zone preferences along the mission}
Simulation Task~2 follows the complete WP5--WP6--WP4--WP3--WP5 mission. Aisle exclusions remove blocked alternatives and zone costs influence the remaining choices. Physical Task~2 connects WP3 to WP4 with costs favoring the corridor containing a small box over the one occupied by another AMR.
\begin{figure}[htbp]
\centering
\includegraphics[width=\linewidth]{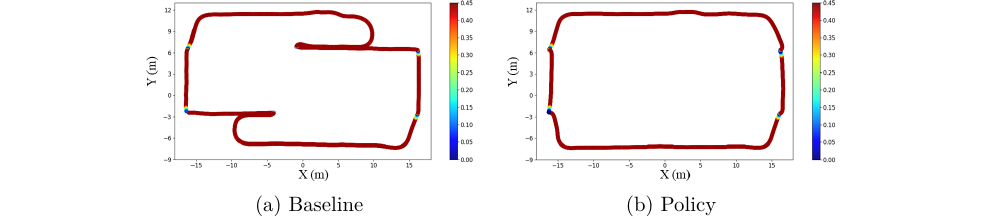}
\caption{Simulation Task~2. Both panels show the complete multi-waypoint mission.}
\end{figure}
\begin{figure}[htbp]
\centering
\includegraphics[width=\linewidth]{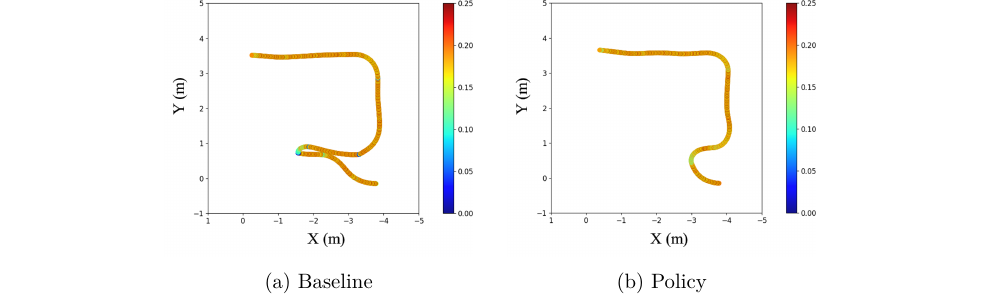}
\caption{Physical Task~2. Zone costs change the selected corridor.}
\end{figure}

\FloatBarrier
\subsection{Person-conditioned avoidance}
Task~3 connects WP7 to WP8 in simulation and WP5 to WP6 in the physical maze. The policy adapter adds a region around the detected person. The traces show the resulting route changes.
\begin{figure}[htbp]
\centering
\includegraphics[width=\linewidth]{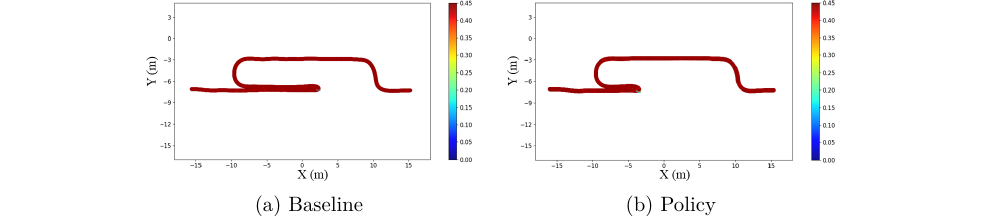}
\caption{Simulation Task~3 with the person-dependent cost region.}
\end{figure}
\begin{figure}[htbp]
\centering
\includegraphics[width=\linewidth]{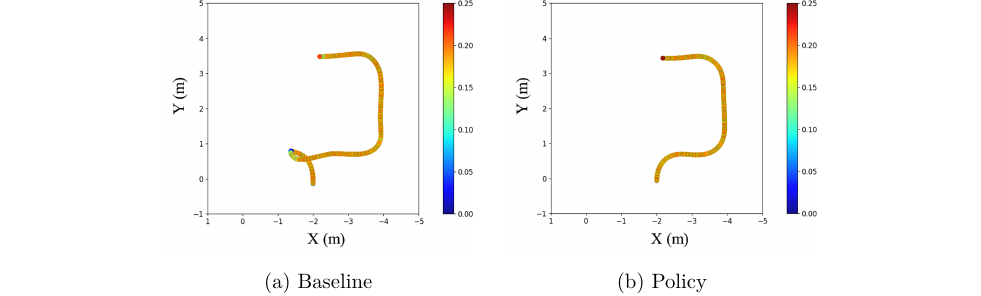}
\caption{Physical Task~3. The policy changes the route taken through the maze.}
\end{figure}

\clearpage
\subsection{Event permission in the warehouse}
Simulation Task~4 connects WP1 to WP2. Both conditions use the policy-enabled stack. Priority-On activates the passage permission and removes the applicable policy prohibition, with sensor obstacles still active.
\begin{figure}[htbp]
\centering
\includegraphics[width=\linewidth]{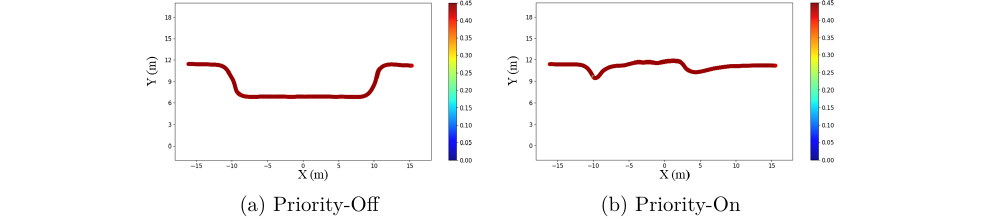}
\caption{Simulation Task~4. This event-permission comparison is distinct from Baseline versus Policy.}
\end{figure}

\FloatBarrier
\section{Shared policies in simulation}
\label{supp:sharing}
\begin{figure}[htbp]
\centering
\includegraphics[width=\linewidth]{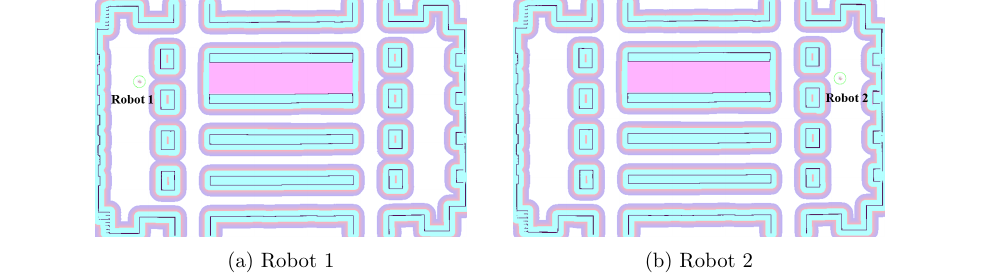}
\caption{Shared policy projection on two simulated robots. The policy contribution agrees even when local sensor information and planned paths differ.}
\label{fig:sim-sharing}
\end{figure}

Two simulated DeliveryRobots receive the same policy-derived updates. The overlays in Figure~\ref{fig:sim-sharing} show matching constrained regions, complementing the physical two-robot views in Figure~\ref{fig:sharing}. Each robot uses an independent Nav2 instance and its own pose and sensor costs.

\section{Resources and reproducibility}
\label{supp:resources}
\subsection{Inference workload}
The resource profile uses the language benchmark workstation with Ollama~0.32.1 and a Vulkan backend. It measures policy generation only, excluding Nav2 and Gazebo. GPU utilization and VRAM are mean increases relative to a model-unloaded 30-second baseline immediately preceding each model. CPU utilization and resident set size (RSS) are means over the Ollama process tree.
\begin{table}[htbp]
\centering\small
\caption{Local-inference resources. GPU change is in percentage points (pp). Host memory is RSS.}
\label{tab:resources}
\begin{tabular}{@{}lrrrr@{}}\toprule
Model & GPU $\Delta$ (pp) & VRAM $\Delta$ (MiB) & CPU (\%) & Host RSS (GiB) \\\midrule
Qwen2.5-7B & 54.7 & 4547 & 16.8 & 0.590 \\
Qwen3-8B & 58.3 & 5319 & 15.6 & 0.609 \\
Qwen3.5-9B & 58.7 & 6438 & 17.4 & 1.005 \\
Llama 3.1-8B & 61.8 & 5043 & 12.8 & 0.567 \\\bottomrule
\end{tabular}
\end{table}

\FloatBarrier
\subsection{Code and benchmark data}
The implementation tested in Section~\ref{supp:checks} accepts integer soft costs from 0 to 252. The preserved language-evaluation snapshot accepts 0--253 and applies schema and within-policy conflict checks. The implementation checks additionally cover active mission restrictions. Neither the 150 reference policies nor the 6,250 stored outputs contain soft cost 253. Changing only the archived validator's upper bound to 252 leaves its acceptance decisions and rejection reasons unchanged for all stored outputs.

The benchmark contains 150 conversion commands with reference signatures and canonical policies, 100 contradictory commands, and 15 demonstrations. The public evaluation bundle is available under \code{benchmarks/evaluation} at commit \code{df3bb05}, including datasets, stored responses, source snapshots, and scoring results.

The language runs record model tags \code{llama3.1:latest}, \code{qwen2.5:7b}, \code{qwen3:8b}, and \code{qwen3.5:9b}, with temperature 0, \code{top\_p}=0, seed 42, \code{repeat\_penalty}=1, \code{num\_predict}=512, \code{think=false}, and JSON output. The 15-demonstration prompt and dataset identifiers are recorded in the run manifest. The separate resource profile uses Q4\_K\_M artifacts.

The public evaluation bundle includes parsed model outputs, references, recorded validation decisions, and offline scoring scripts. Native/custom records include ten run-level results with action status and odometry; Tasks~1--4 retain the reported navigation summaries described in Section~\ref{supp:statistics}. The \href{\ProjectURL}{project page} provides access to the implementation, benchmark data, and demonstrations.

\ifanonymous\else\subsection{Author contact information}
\begingroup\small
\noindent Jiyeong Chae: \href{mailto:cowldud3@dgist.ac.kr}{cowldud3@dgist.ac.kr}\\
Hyunkyo Seo: \href{mailto:hysun43@dgist.ac.kr}{hysun43@dgist.ac.kr}\\
Heonjae Lee: \href{mailto:leehj0806@dgist.ac.kr}{leehj0806@dgist.ac.kr}\\
Kyung-Joon Park: \href{mailto:kjp@dgist.ac.kr}{kjp@dgist.ac.kr}
\endgroup
\fi

\clearpage
\section{Policy scoring and accepted-error analysis}
\label{supp:scoring}
\subsection{Raw and post-validation policy accuracy}
Raw accuracy compares the generated policy with the reference before validation. The comparison retains all destinations and transit entries, including additional coordinate goals and malformed objects. A policy is incorrect if it omits a required element or adds or changes an executable element. Relative restriction durations are compared in minutes.

Post-V compares the complete normalized policy with the reference after acceptance. Destination order is preserved; unordered rule lists are compared independently of serialization order. Both scores check destinations, transit zones, static and timed restrictions, zone costs, object radii, and event conditions, actions, and priorities. Relative durations and radii use three-decimal precision. Interpreter and validator decisions remain those recorded during the experiment.

The comparison covers all 3,750 stored results: five methods, 150 commands, and five calls per command. Per-command outcomes agree across the five repetitions. Llama~3.1-8B has 106 raw matches and 107 correct acceptances; Qwen2.5-7B has 122 at both stages. The accompanying scoring script and case records provide the command-level results.

\subsection{Qwen2.5-7B: malformed transit entry}
Command \code{N03-L0} requests: ``Go to WP3 with traversal cost 5 for A and 200 for B.'' The output contains the requested destination and costs but adds the following transit entry:
\begin{quote}\small\ttfamily
"via\_zones": [\{"dest": \{"x": 0, "y": 0, "yaw": 0\}\}]
\end{quote}
The reference has an empty \code{via\_zones} list. The candidate is incorrect before validation and is rejected with \code{field 'via\_zones' contains a non-string zone}. This outcome occurs in all five repetitions. The rejection concerns zone parsing; the requested costs, 5 and 200, fall within both source versions' cost ranges.

\subsection{Llama 3.1-8B: additional and repeated destinations}
For \code{N09-L4-V3}, the instruction requests WP5 through B, a keepout in C, cost 40 in A, and a 2~m person-avoidance radius. The accepted policy appends an unrequested coordinate goal at $(0,0,0)$ after WP5. It is incorrect at both scoring stages. In \code{N09-L3-V3}, an additional destination with null coordinates is also incorrect before validation and is rejected by the validator.

In \code{N06-L4}, the raw output repeats WP2. Validation removes the duplicate while preserving the requested 30-minute restriction, producing a correct normalized policy. This normalization accounts for Llama's increase from 106 raw matches to 107 correct acceptances. The three outcomes are consistent across all five repetitions.

\clearpage
\subsection{Error types and potential operational effects}
\label{supp:error-types}
The four LLMs produce 40 incorrectly accepted model--command cases under full-policy comparison. Their proportions among accepted policies are 20/127 (15.7\%) for Llama~3.1-8B, 10/132 (7.6\%) for Qwen2.5-7B, 7/148 (4.7\%) for Qwen3-8B, and 3/145 (2.1\%) for Qwen3.5-9B. Table~\ref{tab:error-types} classifies differences between the accepted policies and their references. Multiple labels may apply to one case; the five repeated calls do not increase the number of distinct model--command cases.

\begin{table}[htbp]
\centering\small
\caption{Accepted-error types in stored LLM outputs. Columns follow the model order used in Table~\ref{tab:conversion}; types overlap.}
\label{tab:error-types}
\begin{tabularx}{\linewidth}{@{}Xrrrr@{}}\toprule
Type & Llama & Qwen2.5 & Qwen3 & Qwen3.5 \\\midrule
D: destination or visit-order error & 7 & 2 & 2 & 1 \\
R: requested keepout or person-avoidance omission & 2 & 1 & 2 & 1 \\
N: cost omission, target, or role error & 3 & 2 & 1 & 0 \\
E: event condition, action, or priority error & 0 & 0 & 0 & 0 \\
A: unrequested transit or avoidance constraint & 14 & 7 & 3 & 1 \\\midrule
Distinct incorrect acceptances & 20 & 10 & 7 & 3 \\\bottomrule
\end{tabularx}
\end{table}

\paragraph{Required constraints.}
Six cases omit a requested keepout or person-avoidance rule. Such an omission can leave an intended restriction absent from the planner's costmap unless another layer supplies it. Missing a baseline keepout in an event-conditioned instruction is classified as R; type E concerns the event rule itself. These six cases warrant particular review because they weaken the requested operating constraints.

\paragraph{Destinations and added constraints.}
Destination errors can omit a required stop, change the visit order, or introduce an additional goal. Added transit or avoidance constraints can force detours or make a route unavailable. Cost-role errors include interpreting a cost of 40 as a 40-minute prohibition; the consequences depend on map geometry, active restrictions, and layer combination.

\begin{table}[htbp]
\centering\small
\caption{All three Qwen3.5-9B incorrect acceptances and their potential effects, inferred from the recorded policy content.}
\label{tab:qwen-errors}
\begin{tabularx}{\linewidth}{@{}lXX@{}}\toprule
Command & Policy difference & Potential operational effect \\\midrule
\code{N01-L1-V2} & Adds person avoidance with radius 1.5~m. & Additional avoidance may alter route choice. \\
\code{N02-L1-V2} & Keeps WP7 but omits the required preceding WP2 stop. & The requested pickup or visit may be skipped. \\
\code{N09-L3-V2} & Omits the C keepout while retaining transit B, cost 40 in A, and person radius 2~m. & Zone C may remain available to the planner. \\\bottomrule
\end{tabularx}
\end{table}

These are policy-content analyses of potential effects, not measured incident or constraint-violation rates. They motivate instruction--policy previews, confirmation of critical restriction changes, clarification of numeric roles, and authorization checks for persistent restrictions. These measures are proposed extensions; their effectiveness requires separate evaluation.

\FloatBarrier
\end{document}